\documentclass[runningheads]{llncs}

\usepackage[T1]{fontenc}
\usepackage{booktabs}
\usepackage{graphicx}
\usepackage{color}
\begin{document}
\title{Achieving Scalable Robot Autonomy\\ via neurosymbolic planning using \\lightweight local LLM}
\titlerunning{Achieving Scalable Robot Autonomy via neurosymbolic planning}
%
\author{Nicholas Attolino\inst{1} \and
Alessio Capitanelli\inst{2}\orcidID{0000-0001-8495-5987} \and
\\Fulvio Mastrogiovanni\inst{1}\orcidID{0000-0001-5913-1898}}
\authorrunning{N. Attolino et al.}
%
\institute{Department of Informatics, Bioengineering, Robotics and Systems Engineering, University of Genoa, Genoa, Italy
\email{fulvio.mastrogiovanni@unige.it}\\
\url{https://www.dibris.unige.it} \and
Teseo Srl, P.zza Nicolò Montano, 16121 Genoa, Italy\\
\email{alessio.capitanelli@teseotech.com}\\
\url{https://www.teseo.tech}}
\maketitle              
\begin{abstract}
    PDDL-based symbolic task planning remains pivotal for robot autonomy yet struggles with dynamic human-robot collaboration due to scalability, re-planning demands, and delayed plan availability. Although a few neurosymbolic frameworks have previously leveraged LLMs such as GPT-3 to address these challenges, reliance on closed-source, remote models with limited context introduced critical constraints: third-party dependency, inconsistent response times, restricted plan length and complexity, and multi-domain scalability issues. We present Gideon, a novel framework that enables the transition to modern, smaller, local LLMs with extended context length. Gideon integrates a novel problem generator to systematically generate large-scale datasets of realistic domain-problem-plan tuples for any domain, and adapts neurosymbolic planning for local LLMs, enabling on-device execution and extended context for multi-domain support. Preliminary experiments in single-domain scenarios performed on Qwen-2.5 1.5B and trained on 8k-32k samples, demonstrate a valid plan percentage of 66.1\% (32k model) and show that the figure can be further scaled through additional data. Multi-domain tests on 16k samples yield an even higher 70.6\% planning validity rate, proving extensibility across domains and signaling that data variety can have a positive effect on learning efficiency. Although long-horizon planning and reduced model size make Gideon training much less efficient than baseline models based on larger LLMs, the results are still significant considering that the trained model is about 120x smaller than baseline and that significant advantages can be achieved in inference efficiency, scalability, and multi-domain adaptability, all critical factors in human-robot collaboration. Training inefficiency can be mitigated by Gideon's streamlined data generation pipeline.

\end{abstract}
\section{Introduction}
Human-robot collaboration (HRC) is increasingly critical in domains ranging from assistive and service robotics to industrial automation \cite{heyer2010human}, requiring robots to operate dynamically alongside humans while adapting to unpredictable environmental changes \cite{Darvishetal2018}. A key challenge lies in ensuring interaction \textit{fluency} \cite{hoffman2019evaluating}---minimizing delays between human actions and robot responses---to maintain coordination efficiency \cite{Carfietal2019}. Traditional symbolic task planners based on Planning Domain Definition Language (PDDL) face inherent limitations in this context due to their computational complexity \cite{Garrettetal2020}, especially when frequent re-planning is necessary.

Neurosymbolic frameworks such as Teriyaki \cite{Capitanelli_2024} and Plansformer \cite{pallagani2022plansformergeneratingsymbolicplans} have demonstrated promising advances by training Large Language Models (LLMs) to solve symbolic planning problems. Teriyaki, in particular, has achieved near-perfect plan validity on complex single-domain planning tasks. While end-to-end planning times remain longer than traditional planners, concurrent planning-execution workflows can significantly reduce waiting times by exploiting LLMs' ability to stream plans incrementally as they are generated, rather than requiring a complete solution upfront as traditional planners do.

Nevertheless, two major limitations characterize these approaches: (i) difficulty in scaling across domains, stemming from both the limited context length of earlier LLMs and challenges in assembling large multi-domain datasets; and (ii) reliance on large closed-source cloud-based models for complex tasks, which introduces critical dependencies on third-party systems and variable latency.

These constraints undermine practical deployment in real-world HRC scenarios, particularly in industrial settings where predictable service availability, consistent response times, extended plan horizons, and adaptability across diverse domains are essential.

To address these limitations, we present a novel framework, Gideon, which implements three key contributions:

\begin{enumerate} 
\item A modular pipeline integrating an automated \textit{problem generator} capable of synthesizing large-scale, multi-domain, domain-problem-plan tuples compatible with PDDL 2.1 specifications; 
\item A transition from cloud-based LLMs (e.g., GPT-3) to modern small-scale locally deployable models (e.g., Qwen-2.5 variants), which can be efficiently served over local networks using standard tools such as \textit{text-generation-webui}\footnote{\url{https://github.com/oobabooga/text-generation-webui}}; 
\item Utilization of extended context windows within locally deployed models to enhance scalability toward more complex domains and longer plans. 
\end{enumerate}

The first contribution addresses a significant gap in the field: to our knowledge, no general-purpose tool exists for generating large quantities of random yet realistic and non-trivial problem instances across arbitrary PDDL domains. Currently, only domain-specific generators, such as those used in the International Planning Competitions (IPC), are readily available\footnote{\url{https://github.com/AI-Planning/pddl-generators}}. This technical limitation has hindered the assembly of large-scale datasets spanning multiple PDDL domains. Thus, Gideon's problem generator module could prove extremely valuable for advancing research on generative AI models for task planning.

Regarding the second contribution, by decoupling from proprietary APIs and optimizing model architectures for edge deployment, Gideon avoids latency and availability issues related to the use of third-party cloud services, potentially reduces training and operational costs, and fosters models reuse and incremental improvement.

To evaluate Gideon's efficacy, we adopt both Teriyaki \cite{Capitanelli_2024} and a traditional planner, Probe \cite{lipovetzky2011searching}, as baselines. Our intent is not to surpass the performance of these baselines but to assess the feasibility and scalability of our approach. It must be noted that comparison with Teriyaki is only possible with respect to published results, as GPT-3, the cloud-hosted LLM it was based on, has become inaccessible due to being discontinued. This event highlights even more the importance to develop new neurosymbolic planners based on open source base models.

We employed Gideon to train three single-domain models on one of the benchmark domains presented in \cite{capitanelli2018manipulation}, which was also used in Teriyaki. This domain involves robot manipulation of an articulated object and presents significant challenges for both LLM and traditional planners due to conditional effects that propagate joint angle changes downstream. The three models differ in training dataset size (8k, 16k, and 32k data points) to assess scalability with respect to available training data. Additionally, to provide a preliminary proof-of-concept for multi-domain scenarios, we trained a fourth model (16k) on two different variants of the articulated object manipulation domain.

All models were trained for 4 epochs using Qwen-2.5 1.5B \cite{qwen2.5} as the foundation model, which contains approximately 1.5 billion parameters—two orders of magnitude fewer than the 175 billion parameters of GPT-3, the base model used in the baseline approach.

Our evaluations demonstrate:

\begin{itemize} 
\item That high percentages of valid plans can be achieved also by smaller LLM at the cost of a reduced learning efficiency, as shown by the 32k single-domain model, which generates 66.1\% valid plans on a test set of 1,000 problem instances;
\item A clear path to scaling the percentage of valid plans through additional training data, as evidenced by the 75\% performance increase between the 16k and 32k single-domain models;
\item That multi-domain training is not only feasible, but that it can have a positive impact on training efficiency, although this effect must be confirmed through further investigation.
\end{itemize}

More in detail, a qualitative analysis of the plans generated by the single-domain models, shows that the combination of reduced model size and longer context windows affects the model's ability to maintain long-term coherence. This has a significant impact on training efficiency as, in a single-domain comparison between Teriyaki and Gideon, the former achieved a planning validity ratio of approximately 60\% after training with just 2k samples over 2 epochs, compared to the 32k samples over 4 epochs of the latter. Nevertheless, results highlight a favorable trade-off between model size and data requirements, and the effect seems to be mitigated by data variety in multi-domain training. Considering these factors and how Gideon's problem generator streamlines the generation of additional data, we consider the use of lightweight local LLMs to solve complex logical tasks a research direction worthy of further exploration.

The remainder of this paper details Related Work (II), Methodology (III), Experimental Validation (IV), Limitations (V), and Conclusions (VI). Code and generated data are available under open licenses\footnote{Gideon code available at \url{https://github.com/NichAttGH/Gideon}}.

\section{Related work}

Symbolic task planning is a core subfield of Artificial Intelligence focused on generating sequences of actions to achieve specific goals within defined environments. Traditional methods rely on structured representations like the Planning Domain Definition Language (PDDL), which enables the definition of planning domains and problems in terms of objects, predicates between objects, and actions characterized by preconditions and effects, also expressed as predicates \cite{article_pddl}. In this context, a plan is a sequence of actions that transitions from an initial state to a desired goal state, both expressed as sets of predicates. Symbolic task planning has been widely adopted to enhance robot autonomy due to its explainability and straightforward integration into engineered robot architectures.

However, traditional planners use search algorithms to explore possible states of the modeled environment, and typically exhibit combinatorial explosion in both time and space complexity as the number of symbols increases \cite{Garrettetal2020}. This limitation restricts the applicability of traditional symbolic planners in real-world, dynamic human-robot collaboration scenarios where frequent re-planning is needed due to changing circumstances.

Traditional solutions to this issue, such as domain-specific heuristics and hierarchical planning strategies \cite{Mastrogiovannietal2004,Muralietal2020,Darvishetal2021}, can only mitigate these problems and often require careful engineering.

Neurosymbolic approaches are an alternative class of solvers that also require careful engineering at current state of the art, but hold the promise to to address these limitations. They encode knowledge in vector-based representations for efficient learning from data, while making discrete symbols available by extracting knowledge from trained networks \cite{garcez2020neurosymbolic}.  Such approaches frequently leverage Large Language Models (LLMs)—machine learning architectures trained to read, understand, and generate human-readable text, capable of answering complex questions, solving reasoning tasks, and even producing functioning code \cite{chen2021evaluating}. Initially, popular LLMs were very large, closed-source models available only through APIs, such as OpenAI's GPT-3 \cite{brown2020language} with 175 billion parameters. More recently, smaller open-source LLMs have gained traction due to their favorable speed/performance ratio and the possibility for practical on-premise hosting. This category was pioneered by models like Meta's LLama 2 family \cite{touvron2023llama}, which includes competent models with as few as 7 billion parameters. Today, thanks to advancements in model architecture and training data quality, models like Microsoft Phi-3.5 \cite{abdin2024phi}, with only 3.8 billion parameters, can outperform GPT-3 on important logical benchmarks like MATH \cite{hendrycks2021measuring} and the ARC challenge, demonstrating the potential of smaller models in complex reasoning tasks.

A recent comprehensive review \cite{Pallagani_2024} identified eight main aspects of Automated Planning and Scheduling that LLMs can contribute to, namely: \textit{Language Translation}, \textit{Plan Generation}, \textit{Model Construction}, \textit{Multi-Agent Planning}, \textit{Interactive Planning}, \textit{Heuristic Optimization}, and \textit{Tool Integration}. Among these, \textit{Plan Generation} is particularly relevant to our work, focusing on direct generation of plans or decision-making sequences using LLMs.

Techniques in this category can be broadly classified into two approaches: prompting-based strategies \cite{song2023llmplannerfewshotgroundedplanning,lu2023neurosymbolic,wang2023planandsolvepromptingimprovingzeroshot} and fine-tuning-based approaches \cite{pallagani2022plansformergeneratingsymbolicplans,Rossetti_Tummolo_Gerevini_Putelli_Serina_Chiari_Olivato_2024,Ghafarian_Tamizi_2024,Capitanelli_2024}. The former provides specific instructions (i.e., a \textit{prompt}) to an unchanged model, while the latter requires additional training to condition the model to express specific behaviors. Other significant approaches include self-verification mechanisms \cite{valmeekam2022large}, hierarchical planning \cite{10610434}, and integration with traditional planners \cite{Ding_2023}.

Prompt-based approaches often struggle to accurately map admissible actions, as even a single incorrect token can invalidate a PDDL plan that is otherwise \textit{linguistically} coherent to the model's eyes. Consequently, they typically yield lower percentages of valid plans compared to fine-tuning-based approaches. Given our focus on enabling reliable and efficient operation in real-world HRC scenarios, our approach is based on the results of the fine-tuning-based methods.

The Teriyaki framework \cite{Capitanelli_2024} has demonstrated two important capabilities: training LLMs to generate long and complex PDDL plans with high accuracy (95.5\% in single-domain applications) and exploiting LLM characteristics to generate plans action-by-action, significantly reducing waiting times for a plan. This incremental plan generation strategy is highly desirable in HRC scenarios and leads us to exclude methods that reintroduce end-to-end planning, such as those that achieve high accuracy through integration with traditional planners.

Nevertheless, the Teriyaki framework and many other approaches still face significant limitations: many rely on large cloud-based and now outdated or discontinued models such as GPT-3, and they struggle with generalization across different domains due to factors like data availability and limited context length. 

Regarding data availability, to our knowledge there are no tools or frameworks capable of generating large amount of unique and realistic problems instances given a domain, even though recent works have shown progress on the automatic domain generation side \cite{khandelwal2024pddlfuse,tantakoun2025llms}.

Regarding context length limitations, it is useful to consider that Teriyaki trained single-domain models using exclusively problem-plan pairs, as including the domain would have consumed too much of GPT-3's limited context window (i.e., 2048 tokens or about 8000 characters) given the expected plan lengths.

These limitations highlight the need for a new neurosymbolic framework that can overcome these constraints. We argue that local lightweight LLMs offer a promising solution, given their recent success on logical reasoning tasks and the extended context windows of newer models. Such a framework would enable on-device execution, support multi-domain planning, and provide more predictable response times while maintaining the benefits of neurosymbolic approaches in terms of efficiency and adaptability. This direction forms the foundation of our proposed Gideon framework, which addresses these limitations through innovative approaches to data generation and support for local LLMs.

\section{Methodology}

\subsection{Planning domains, PDDL version and planner}
For our evaluation, we selected two PDDL domains referred to as MACRO and NO-MACRO, both modeling manipulation actions on articulated objects. These domains have been previously used in human-robot collaboration scenarios and as benchmarks for Teriyaki, enabling direct comparison with baseline models.

The two models differs mainly in their action set, where the former operates at a higher level of abstraction than the latter. Both domains utilize an \textit{absolute representation}, where angles between pairwise links of the articulated object are expressed relative to an absolute reference frame. This formulation presents significant challenges for both traditional planners and LLMs. Traditional planners must propagate conditional effects to modify state variables not directly affected by actions, while LLMs struggle to maintain plan coherence due to these conditional effects. A complete analysis of the two domains is outside the scope of this work, but details about their formulation and performance, both with traditional and neurosymbolic planners, can be found in literature \cite{capitanelli2018manipulation,Capitanelli_2024}.

Since the domains share fundamental similarities, some transfer learning occurs during multi-domain training, so multi-domain results should be viewed primarily as a proof of feasibility. Given that both domains employ conditional effects, we used PDDL 2.1 and the Probe planner \cite{lipovetzky2011searching}, which was also used in previous work by \cite{capitanelli2018manipulation}.

Probe is a satisficing planner that generates action sequences by exploring the search space using a greedy, depth-first approach (a probe). This method often quickly finds solutions, though not necessarily optimal ones. Probe can identify landmarks—key states reached by different probes—which can be used to generate sub-plans between these landmark states.

We selected Probe for two reasons: it allows direct comparison with Teriyaki and its widespread adoption facilitates adaptation of our approach to different scenarios. However, it should be noted that the choice of planner influences result quality, and alternative planners could potentially be used.

\subsection{Training pipeline}

\begin{figure}[h!]
\centering
\includegraphics[width=\columnwidth]{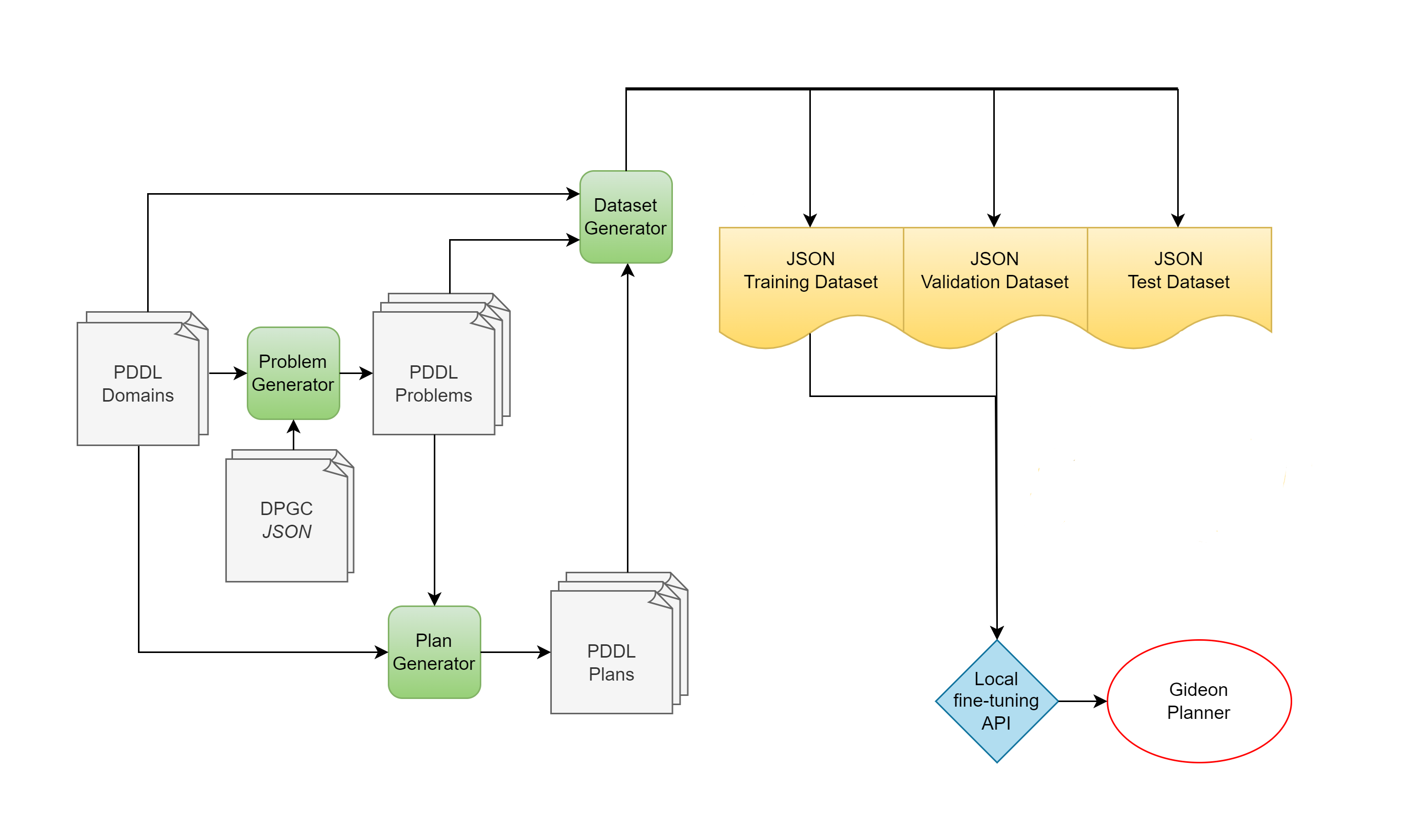}
\caption{A diagram of Gideon's training pipeline. In order, selected domains are processed in accordance to the rules specified in the DPCG file to generate a set amount of problem instance; then plans are generated by the Plan Generator module with the selected planner; finally, domain-problem-plans tuples are assembled and formatted into the training, validation and test set files.}
\label{fig:pipeline}
\end{figure}

The core of the Gideon's framework is the training pipeline depicted in Figure \ref{fig:pipeline}. In order, a General Problem Generator takes as input one or more PDDL domains and their relative Domain-Problem Generation Configuration files (DPGC). Each DPGC specifies in a specific JSON schema the rules to randomly generate problems according to user rules. Then, a Plan Generator module, which can be reconfigured to support several planners, plans against the previously generated domain-problem pairs and validates the resulting plans using the well-known VAL validation utility \cite{howey2004val}. If the planner does not provide plans in the format required by VAL, such as in the case of Probe, a conversion takes place to ensure compatibility. This is of utmost importance to also ensure that the plans that will be generated by the resulting model will follow the same specifications, enabling seamless testing. Finally, the Dataset Generator module assembles appropriate domain-problem-plan tuples, shuffles them, and inserts them into training, validation and test dataset in the Alpaca JSON Training Format. The Alpaca format is a list of JSON tuples, namely: \textit{instruction}, used to provide commands to the model; \textit{input}, used to provide additional information useful for the task at hand, such as data or examples; and \textit{output}, i.e., the expected output of the model. in Gideon, domain, problem and plan are mapped to instruction, input and output respectively. Throughout the whole process, it is guaranteed that all data points are unique, i.e., eventual duplicates are immediately discarded to avoid dataset imbalances and data leakage from the training set to the validation and test sets. 

The specific implementation provided in this paper allows the user to specify the total amount of data point, i.e., domain-problem-plan tuples to be generated, as well as their proportions among the final training, validation and test dataset; it also ensures that the specified quotas are met even if a plan is lost, for example due to planner timeout or a data point being discarded as a duplicate of an existing one. Both the Problem and Plan generation modules allow to easily add data to an existing dataset as well as pause and resume generation, while always ensuring there are no duplicates in the final datasets to avoid unbalances and data leakage. All modules work together to facilitate dataset management by storing intermediate data, datasets and the relative logs in a structured directory.

The Plan Generator module is the only element of the pipeline that might require user customization, but modular design mitigates this limitation. If a different planner is needed, it is possible to add support for a new planner simply by adding a new planner class that points to the planner's executable, correctly mapping input parameters and, if the output format is not-compliant with VAL, implementing an adequate translation method. In this way, a library of supported planners can be progressively built as needed. The Problem Generator can instead be completely configured via a Domain-Problem Generation Configuration file, which we will discuss in the next section.

\subsection{Domain-Problem Generation Configuration}
The generated data in our research is not \textit{synthetic} in a derogatory sense. Since we focus on symbolic task planning with assumed perfect perception and grounding capabilities of the agent, the planner cannot distinguish between generated and real-world data. Therefore, we exclusively use generated data for training and evaluation, as it provides equivalent validation for the planning module while eliminating variability from perception and execution components.

Nevertheless, data generation for AI planning domains requires strategic approaches rather than random instantiation of objects and predicates. Random generation could produce unsolvable or overly complex problems. While strict realism isn't necessary—indeed, abstract domains that exercise logical reasoning abilities are valuable—data should align with each domain's \textit{intended use}.

This means ensuring that (i) problems are initialized in states from which goals are reachable; and, more generally, (ii) certain domain-specific rules implied by the domain designer are respected, even if not encoded in the domain itself.

Both points are related to the fact that predicates often have intended usage patterns. Regarding the first point, in the domains considered here for example, articulated objects must be grasped either with both robotic grippers or none. Initializing with only one gripper grasping would lead to unsolvable problems.

Regarding the second point, often certain predicates in the problem serve as static encoders of general properties. For example, the domains considered in this work do not represent joint angle configurations (e.g., a joint at 45 \textit{deg} can rotate to 60 \textit{deg} clockwise or 30 \textit{deg} counter-clockwise in 15 \textit{deg} increments). That specification is instead encoded in the problem, thus it should be consistent across problems, or at least, modifications should be controlled by the user with care in order to avoid unintended behaviors.

To address the challenges in generating appropriate planning problems, we introduce the Domain-Problem Generation Configuration (DPGC) format. The schema is based on JSON and allows to express fine-grained control over problem generation offering a principled solution to both concerns: ensuring problem solvability and respecting domain-specific rules. An overly detailed description of DPGC format is outside the scope of this work. Below is an overview of the format main concepts and features, while both the JSON schema and a practical example are available on the project's repository\footnote{\url{https://github.com/NichAttGH/Gideon/tree/main/Gideon/jsons}}.

DPGC tackles problem generation in a hierarchical and modular way, by decomposing the different constituting parts of a generic PDDL problem into subsets of predicates that can be then randomly instantiated according to specific rules. At the highest level, the user can specify a \textit{constant initial state} component, a \textit{variable initial state} component, and a \textit{variable final state}. The constant initial state specifications are inserted verbatim into problems, ensuring critical invariants remain consistent across all generated problems. 

The variable states are instead defined as a list of \textit{predicate pools}, which in turn are defined as list of predicates. For each predicate it is possible to specify the probability that it will be generated, how many instances of the predicate should be generated, and an ordered list of arguments. The arguments are expressed as \textit{object pools}, where each pool specifies a type, quantity of items in the pool, naming convention, and usage mode, such as \textit{mutex} (objects can be selected from the pool only once) or \textit{sequential} (objects must be used in sequence rather than randomly). When a predicate is generated, it is instantiated with object arguments from the specified pool and according to its usage mode. If no usage mode is specified, a random object from the pool is retrieved. This approach ensures consistent object generation across problems while allowing for controlled variations. Predicate pools in variable states can also be mutexed among themselves to allow for specific sub-configurations to be exclusively selected with a desired probability (e.g., whether an object is grasped with both grippers or instead grippers should be both marked as free).

Additionally, a synchronization mechanism using tagged references ensures logical relationships among object pools in a predicate pool. When multiple predicates must maintain consistency, the tag ensures that related objects are selected together or in a specific order to establish relationships (e.g., \texttt{link-pool\$0} and \texttt{link-pool\$0+1}). For example, we might want to generate the predicates \texttt{(in-hand link1)} and \texttt{(in-hand link2)} while ensuring the two links are always adjacent. In this case, only the first predicate will be initialized with a random object from the relative pool, while the second will be selected in sequence.

Using these basic elements in combination, DPGC provides the expressive power and flexibility to be applied to a wide range of PDDL domains. Currently the tool supports all the fundamental and most common requirements of PDDL 2.1. A notable exception are numeric fluents, which are planned but not yet implemented.

\subsection{Training and testing}

We employed Qwen-2.5 (1.5B parameters variant) \cite{qwen2.5} as our foundation model, selecting it for its balance between compactness and capability. The model's small size enabled full fine-tuning rather than parameter-efficient methods like LoRA, which often underperform in logical reasoning tasks \cite{biderman2024lora}. We also utilized the half-precision floating-point format (\textit{fp16}) to maintain model fidelity.

Our experimentation comprised four distinct training configurations. First, we developed three single-domain models using increasing quantities of NO-MACRO domain data: 8.000, 16.000, and 32.000 samples (i.e., domain-problem-plan tuples), resulting in models designated as Gideon-NO-MACRO-8k, Gideon-NO-MACRO-16k, and Gideon-NO-MACRO-32k respectively. These models serve to benchmark against the Teriyaki baseline and evaluate performance scaling with dataset size. We then trained a multi-domain model, Gideon-MD-16k, using 16,000 total samples, 8,000 each from MACRO and NO-MACRO domains, in order to assess cross-domain generalization capabilities. All training configurations utilized a 1,000-sample validation set, equally divided between domains for the multi-domain model.

A critical consideration in LLM-based planning is that optimization targets linguistic coherence rather than planning validity. This creates a proxy relationship where we leverage linguistic fluency as an indicator of logical soundness. Consequently, training metrics differ from true \textit{planning validity}, which we define as the percentage of plans that are both (i) formally correct regarding domain specifications and preconditions, and (ii) successful in reaching goal states. Post-training evaluation therefore involved testing on 1,000 domain-problem pairs, selected by making sure there was no data leakage between training, validation and test set. Again, samples were equally split between domains for Gideon-MD-16k.

To streamline the training process, we leveraged LLamaFactory \cite{zheng2024llamafactory}, which provides comprehensive training infrastructure with a much more granular training parameter control than the OpanAI platform used by Teriyaki. Key configuration choices included: input sequence cut-off at 3,096 tokens to optimize memory usage; reduced \textit{per device train batch size} from 2 to 1 to prevent memory limitations; and standard hyperparameters including 4 training epochs, $5.0 \times 10^{-5}$ learning rate, \textit{cosine} learning scheduler, and 8 gradient accumulation steps.

During validation, we maintained the 3,096 token limit (combining input and output) and set temperature to $0.01$, a common practice in planning applications to prioritize the model's highest-confidence predictions. For testing, we relied instead on \textit{Text Generation Web UI}\footnote{\url{https://github.com/oobabooga/text-generation-webui}}, which provides not only a graphical interface for loading, managing and querying models, but also efficient inference and model serving over local networks via convenient APIs. Temperature was again set to $0.01$ during testing.

All computational work, including dataset generation, model training, and evaluation, was performed on a system equipped with an Intel Core i9-11900 CPU (2.50 GHz), 64GB DDR4 RAM, and an NVIDIA RTX A6000 GPU with 48GB VRAM.

\section{Results and discussion}
We will now discuss the results of our single- and multi-domain tests on the trained models and use the Probe traditional planner and Teriyaki LLM-based neurosymbolic planner as baselines. Unfortunately, it is impossible to perform new tests on Teriyaki due the base model, GPT-3, which is a very large cloud-based LLM, reaching end of life and becoming unavailable to the public. For this reason, we will refer in this section to the originally published results \cite{Capitanelli_2024}, which were obtained from an equivalent test set of 1.000 samples.

\subsection{Single-domain}

\begin{figure}[h!]
\centering
\includegraphics[width=\columnwidth]{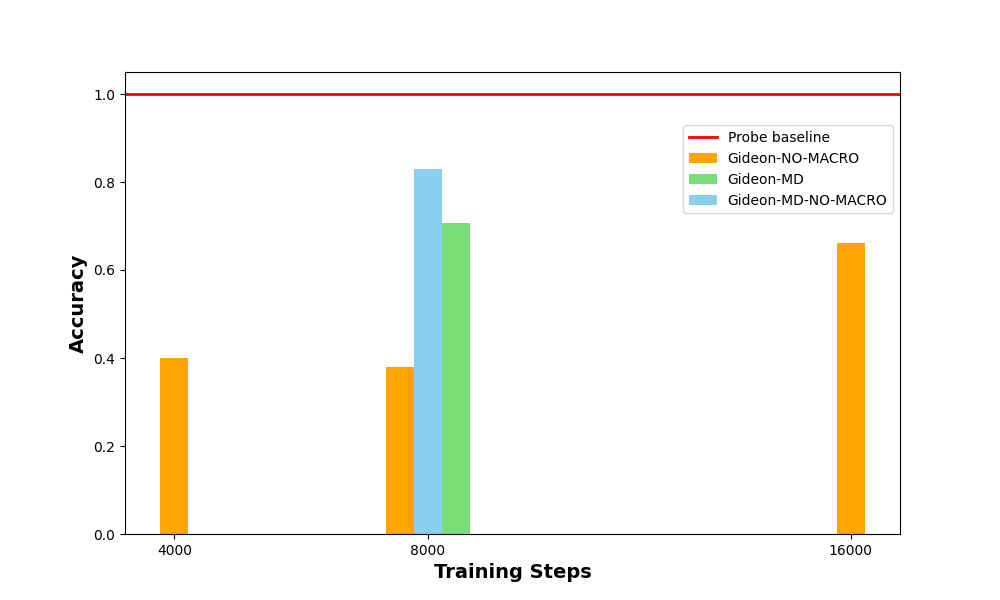}
\caption{Planning validity achieved by Gideon-NO-MACRO and Gideon-MD with respect to increasing training steps. Gideon-NO-MACRO-8k, -16k, and -32k are reported in orange. For Gideon MD, both results on the multi-domain test set (in green) and on a second single-domain test set (NO-MACRO, in blue) are reported to facilitate comparison with single-domain models. The red line above represent the success rate of the baseline traditional planner Probe (i.e., 100\%)}
\label{fig:validity}
\end{figure}

We present the results of our evaluation on three single-domain models: Gideon-NO-MACRO-8k, -16k, and -32k. Figure \ref{fig:validity} reports planning validity reached by the three models, as well the values relative to Gideon-MD-16k both in single and multi-domain testing.

The 8k model demonstrates a respectable $39.9\%$ planning validity rate, while surprisingly the 16k model performs slightly worse at $38.0\%$. The 32k model, however, achieves a substantial improvement with $66.1\%$ validity, representing a $75\%$ increase over the 16k variant. This stagnation followed by an abrupt increase in performance may be attributed to \textit{delayed generalization}, a phenomenon commonly observed when training LLMs on logical sequences, as the model shifts from quick memorization to slow generalization. Another interpretation is the combination of relatively small models processing large datasets containing repetitive action subsequences, potentially causing temporary overfitting before additional training data enables escape. This effect warrants further investigation, as it could impact training efficiency for smaller planning-focused LLMs, which in turn might necessitate more careful hyperparameter tuning and regularization techniques. For comparison, the baseline Teriyaki model trained with GPT-3 achieves comparable results ($60.0\%$) after training on only $2,000$ samples for 2 epochs.

Tables \ref{tab:sd_comparison} and \ref{tab:sd_comparison_time} provide deeper insights into plan quality and computational efficiency. Table \ref{tab:sd_comparison} shows the percentage of valid plans and their lengths (as a proxy for plan quality), while Table \ref{tab:sd_comparison_time} presents timing statistics including average, minimum, maximum, and median planning times with standard deviations across the test set, and with Probe included as baseline.

\begin{table}[h!]
    \centering
    \resizebox{\textwidth}{!}{%
    \begin{tabular}{ccccccccc}
    \toprule
         Solver & Validity (\%) & Avg\_steps & Min\_steps & Max\_steps & Median\_steps \\
    \midrule
         Gideon-NO-MACRO-8k & 39.9 & 37.61 & 17 & 61 & 38 \\
         Gideon-NO-MACRO-16k & 38.0 & 37.06 & 17 & 65 & 37 \\
         Gideon-NO-MACRO-32k & 66.1 & 40.69 & 15 & 68 & 41 \\
    \midrule
         Probe-NO-MACRO & 100 & 35.2 & 2 & 182 & 31 \\
    \bottomrule
    \end{tabular}%
    }
    \vspace{10pt}
    \caption{Comparison of Gideon models against Probe on a single domain (NO-MACRO) on the test dataset.}
    \label{tab:sd_comparison}
\end{table}


Regarding plan quality, Gideon models generate longer plans on average compared to both Probe and Teriyaki. While Teriyaki previously demonstrated a slight advantage over Probe (producing up to $10\%$ shorter plans in some domains), the higher plan lengths in Gideon models align with their lower planning validity rates. Notably, the minimum plan length for Gideon models never falls below 15 actions, whereas Probe solves trivial instances in as few as 2 steps—another potential evidence of overfitting in the Gideon models.

\begin{table}[h!]
    \centering
    \resizebox{\textwidth}{!}{%
    \begin{tabular}{cccccccc}
    \toprule
         Solver & Avg\_t (s) & Min\_t (s) & Max\_t (s) & Median\_t (s) & Std\_t (s) \\
    \midrule
         Gideon-NO-MACRO-8k  & 62.50 & 20.18 & 121.51 & 60.91 & 19.01 \\
         Gideon-NO-MACRO-16k  & 65.61 & 20.05 & 128.08 & 65.52 & 18.53 \\
         Gideon-NO-MACRO-32k  & 64.99 & 18.35 & 127.35 & 65.27 & 18.86 \\
    \midrule
         Probe-NO-MACRO & 3.60 & 0.06 & 37.20 & 2.18 & 4.27 \\
    \bottomrule     
    \end{tabular}%
    }
    \vspace{10pt}
    \caption{Processing times on the same domain (NO-MACRO) for each solver with some statistics.}
    \label{tab:sd_comparison_time}
\end{table}


Despite the smaller model size, end-to-end planning times for Gideon models are significantly longer than both baseline systems, partially explained by their greater average plan lengths. As demonstrated in \cite{Capitanelli_2024}, planning time for LLM-based neurosymbolic planners scales linearly with the combined length of inputs and outputs. Hardware limitations also contribute to these timing differences. Nevertheless, implementing an \textit{action-by-action} planning strategy as presented in \cite{Capitanelli_2024} could potentially reduce planning time for Gideon-NO-MACRO-32k to $1.6s$, a $55\%$ improvement over probe. While inference optimization techniques such as model quantization might offer further improvements, their impact on planning validity requires investigation.

\subsection{Multi-domain}

Our analysis of Gideon-MD-16k reveals competitive planning validity rates across multiple domains as illustrated in Figure \ref{fig:validity}. The model achieves 
70.6\% planning validity on a mixed test set of 1,000 samples evenly distributed between two domains. When evaluated on single-domain test sets of 500 samples each, it reaches 83.0\% validity on the NO-MACRO domain and 58.2\% on the MACRO domain.

Notably, the NO-MACRO domain results surpass those of both Gideon-NO-MACRO-8k (trained on an equivalent number of NO-MACRO samples) and Gideon-NO-MACRO-32k (trained on four times as many samples). This performance improvement may be attributed to a form of \textit{regularization by data variety}, i.e., training across multiple domains compels the model to prioritize generalization over memorization, in this case accelerating the learning process. Qualitative analysis of failed plans indicates that errors typically involve \textit{forgetting} portions of the goal state (e.g., turning the last joint to 15 degrees instead of 315) rather than incorrect action parameterization or domain confusion. These findings suggest that multi-domain planning with lightweight LLMs might be both feasible and scalable.

\begin{table}[h!]
    \centering
    \resizebox{\textwidth}{!}{%
    \begin{tabular}{ccccccccc}
    \toprule
         Solver & Domains & Validity (\%) & Avg\_steps & Min\_steps & Max\_steps & Median\_steps \\
    \midrule
         Gideon-MD-16k & MACRO \& NO-MACRO & 70,6 & 31.26 & 10 & 66 & 29 \\
    \midrule
         Gideon-MD-16k & MACRO & 58.2 & 20.17 & 10 & 38 & 20 \\
         Gideon-MD-16k & NO-MACRO & 83.0 & 39.04 & 13 & 66 & 39 \\
         \midrule
         Gideon-NO-MACRO-8k & NO-MACRO & 39,9 & 37.61 & 17 & 61 & 38 \\
         Gideon-NO-MACRO-16k & NO-MACRO & 38.0 & 37.06 & 17 & 65 & 37 \\
         Gideon-NO-MACRO-32k & NO-MACRO & 66.1 & 40.69 & 15 & 68 & 41 \\
         \midrule
         Probe-NO-MACRO & NO-MACRO & 100 & 35.2 & 2 & 182 & 31 \\
    \bottomrule
    \end{tabular}%
    }
    \vspace{10pt}
    \caption{An analysis of Gideon-MD-16k planning validity rates and plan length metrics. Note that tests of the model on single-domain test sets were performed on only 500 samples. Below, results of the single domain models and the baseline traditional planner Probe are reported for comparison.}
    \label{tab:md_validity}
\end{table}


Table \ref{tab:md_validity} provides more comprehensive plan quality metrics, including domain-specific results for Gideon-MD-16k on the two separate 500-sample test sets. It is important to note that the MACRO domain's expected average plan length is approximately half that of the NO-MACRO domain. In this context, average and median step counts remain consistent with single-domain models. Nevertheless, the reduced minimum number of steps reinforces the hypothesis that the model employs more robust planning strategies. The lower performance on the MACRO domain compared to NO-MACRO remains unexplained, though experiments in literature with Teriyaki also demonstrated poorer performance on NO-MACRO domains in terms of average step count, potentially indicating underlying domain complexities.

\begin{table}[h!]
    \centering
    \resizebox{\textwidth}{!}{%
    \begin{tabular}{cccccccc}
    \toprule
         Solver & Domains & Avg\_t (s) & Min\_t (s) & Max\_t (s) & Median\_t (s) & Std\_t (s) \\
    \midrule
        Gideon-MD-16k & MACRO \& NO-MACRO & 43.97 & 12.37 & 123.24 & 39.9 & 18.27 \\
    \midrule
        Gideon-MD-16k & MACRO & 33.50 & 12.37 & 119.38 & 31.23 & 11.80 \\
        Gideon-MD-16k & NO-MACRO & 54.43 & 14.06 & 123.24 & 52.98 & 17.58 \\
    \midrule
         Gideon-NO-MACRO-8k  & NO-MACRO & 62.50 & 20.18 & 121.51 & 60.91 & 19.01 \\
         Gideon-NO-MACRO-16k  & NO-MACRO & 65.61 & 20.05 & 128.08 & 65.52 & 18.53 \\
         Gideon-NO-MACRO-32k  & NO-MACRO & 64.99 & 18.35 & 127.35 & 65.27 & 18.86 \\
    \midrule
        Probe-NO-MACRO & NO-MACRO & 3.60 & 0.06 & 37.20 & 2.18 & 4.27 \\
    \bottomrule     
    \end{tabular}%
    }
    \vspace{10pt}
    \caption{An analysis of Gideon-MD-16k planning times. Note that tests of the model on single-domain test sets were performed on only 500 samples. Below, results of the single domain models and the baseline traditional planner Probe are reported for comparison.}
    \label{tab:md_times}
\end{table}

\begin{figure}[h!]
    \centering
    \includegraphics[width=\columnwidth]{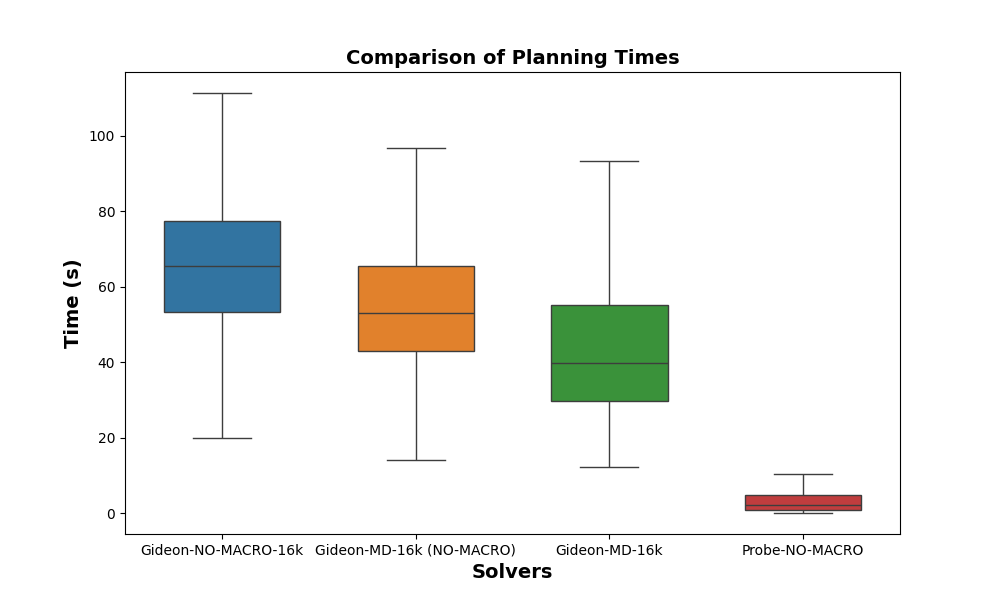}
    \caption{Box plots of the planning times of Gideon-NO-MACRO-16k, Gideon-MD-16k tested on 500 NO-MACRO samples, Gideon-MD-16k tested on the complete multi-domain test set, and the baseline traditional planner Probe.}
    \label{fig:boxplots}
\end{figure}

Table \ref{tab:md_times} and Figure \ref{fig:boxplots} report more detailed results on planning times. Again, it must be noted that plan length for the MACRO domain is expected to be shorter and that planning times for LLM-based neurosymbolic planners depend on plan length. Nevertheless, in this case it is possible to observe an improvement moving from Gideon-NO-MACRO model to Gideon-MD-16k. 

\section{Conclusion}

In this paper, we presented Gideon, a novel neurosymbolic framework designed to address key limitations in existing approaches to human-robot collaboration planning. Our work was motivated by the critical need for interaction fluency in dynamic environments while overcoming the constraints of traditional symbolic planners and cloud-based neurosymbolic solutions.

At its core, Gideon introduces three major innovations: (1) a modular pipeline featuring an automated problem generator capable of synthesizing large-scale, multi-domain datasets compatible with PDDL 2.1; (2) a transition from cloud-based to small-scale locally deployable language models; and (3) the utilization of extended context windows to enhance planning complexity and scalability.

Methodologically, our framework leverages the Qwen-2.5 1.5B parameter model, which is significantly smaller than the 175B parameter GPT-3 used in previous approaches, alongside a carefully designed Domain-Problem Generation Configuration format, to create structured datasets spanning multiple planning domains. Through systematic experimentation with single and multi-domain models trained on varying dataset sizes, we provided preliminary evidence that multi-domain task planning might be feasible and scalable using lightweight, local LLM.

Our results reveal several important findings. First, our 32k single-domain model achieved 66.1\% planning validity, showing that smaller LLMs can effectively generate valid plans when provided with sufficient additional training data. Second, we observed preliminary evidence that multi-domain training can lead to potential benefits both in terms of performance and training efficiency in lightweight LLM, we hypothesize, due to an intrinsic regularization effect. As a consequence, our multi-domain model reaches 83.0\% validity on single-domain testing, outperforming all models trained on a single domain. This phenomenon suggests that exposure to multiple domains helps the model develop more robust planning strategies rather than relying on memorization.

Gideon does face limitations: from a methodological point of view, more tests must be carried out with an increasing number of domains; on the performance side, end-to-end planning times and learning efficiency is inferior compared to larger models. However, our findings suggest these limitations can be mitigated through streaming plan generation strategies and optimization techniques like quantization. More importantly, the framework's data generation capabilities provide a clear and convenient path to improving performance by scaling training data quantity and variety.

The ability to deploy lightweight, locally-hosted models while maintaining reasonable planning validity represents a significant step toward practical neurosymbolic planning, especially in real-world human-robot collaboration scenarios where the possibility to start executing as soon as the first action as been generated can have a significant impact on the interaction fluency. By eliminating reliance on cloud-based services and enabling cross-domain generalization, Gideon offers a more robust, predictable, and adaptable solution for environments where reliable planning is essential.

\begin{credits}
\subsubsection{\ackname} This research is partially supported by the Italian government under the National Recovery and Resilience Plan (NRRP), Mission 4, Component 2, Investment 1.5, funded by the European Union NextGenerationEU programme, and awarded by the Italian Ministry of University and Research, project RAISE, grant agreement no. J33C22001220001.

\subsubsection{\discintname}
The authors have no competing interests to declare that are
relevant to the content of this article. 
\end{credits}
%
%
%
\bibliographystyle{splncs04}
%

\bibliography{biblio}
\end{document}